# Reinforcement Learning based Air Combat Maneuver Generation

Muhammed Murat Özbek[1]

*Department of Aeronautical Engineering, Istanbul Technical University*

Emre Koyuncu[2]

*Department of Aeronautical Engineering, Istanbul Technical University*

## I. Abstract

The advent of artificial intelligence technology paved the way of many researches to be made within air combat sector. Academicians and many other researchers did a research on a prominent research direction called autonomous maneuver decision of UAV. Elaborative researches produced some outcomes, but decisions that include Reinforcement Learning(RL) came out to be more efficient. There have been many researches and experiments done to make an agent reach its target in an optimal way, most prominent are Genetic Algorithm(GA) , A*, RRT and other various optimization techniques have been used. But Reinforcement Learning is the well-known one for its success. In DARPHA Alpha Dogfight Trials, reinforcement learning prevailed against a real veteran F-16 human pilot who was trained by Boeing. This successor model was developed by Heron Systems. After this accomplishment, reinforcement learning bring tremendous attention on itself. In this research we aimed our UAV which has a dubin vehicle dynamic property to move to the target in two-dimensional space in an optimal path using Twin Delayed Deep Deterministic Policy Gradients (TD3) and used in experience replay Hindsight Experience Replay(HER).We did tests on two different environments and used simulations. At first, it is aimed for agent to take the optimal path to the target where there were not any obstacles but only barriers. In each episode our agent started at a random point, our target was stable its position didn't change. It found its own path in an optimal and a fast way. Then in order to test the limits of the mechanism, to make our agent to reach its target harder and to make it perform different maneuvers, we added obstacles. It overperformed and overcome all obstacles. Now the research for making two UAV as multi agents do a dogfight in a two-dimensional space. This research paper proposes an algorithm for motion planning, it's using twin delayed deep deterministic policy gradient (TD3) which is an algorithm tailored for MDP with continuous action that uses reinforcement learning as a base algorithm.

## II. Introduction and Objectives

Improvements in Artificial intelligence paved the way of using new technologies in aviation. One of the prominent inventions is that Unmanned Aerial Vehicles (UAV) are developed and used, they minimize the errors made by pilots. Artificial intelligence and aviation are growing together day by day. Undoubtably the most important aspect is reinforcement learning (RL).

In recent months the most and the best well-known reinforcement real-life application is the one used in aviation by Heron systems. Heron systems has shown outstanding success in Alpha Dogfight Trials Final Event. Those trials were designed with the aim of demonstrating advanced algorithms' capacity of performing simulated movements, within visual range air combat maneuvering, or as known as dogfight. Heron Systems showed a great performance against other competitors in F-16 AI agents, and it dominated them in the main event, in which series of dogfights against a veteran/experienced real human Air Force F-16 pilots. It went 5 to 0 against it, and human pilot was not able to compete against aggressive and precise maneuvers of the AI. It has been once proved again that AI is superior than what humans can achieve.

Reinforcement Learning(RL) is a type of machine learning technique that enables an agent to learn in an interactive environment by trial and error using feedback from its own actions and experiences.

The main difference of reinforcement learning when compared to others is that in reinforcement learning it is not required to have a labelled input and output pairs, instead it focuses on two main things, exploration (exploring unexplored territories) and exploitation (using background knowledge).

[1] E-mail: ozbekm17@itu.edu.tr
[2] E-mail: emre.koyuncu@itu.edu.tr

Reinforcement learning differs from supervised learning in not needing labelled input/output pairs be presented, and in not needing sub-optimal actions to be explicitly corrected. Instead the focus is on finding a balance between exploration (of uncharted territory) and exploitation (of current knowledge)[1]. MDP (Markov decision process) is used to represent the environment, the main reason for this is since most of the reinforcement learning algorithms for this concept use dynamic programming techniques and MDP works best with dynamic programming techniques[2]. The vital difference between the classical dynamic programming and reinforcement learning algorithm is that RL algorithm do not take into consideration of exact model of the MDP when calculated mathematically, and they target for the MDPs in which methods are not feasible. Since it can be generalized RL is worked on by many different disciplines like game theory, operations research, information theory, simulation-based optimization, multi-agent systems and statistics, those are the main areas but there are also many in disciplinary works.

Environment is defined as set of states that "agent" attempts influence if a choice of actions. The environment is modelled with stochastic finite state machine with inputs and outputs, respectively input stands for actions sent from the agent and outputs stand for observations and rewards sent to the agent. In RL the agent decides what action it needs to take. To make this decision it is allows to use any observations made through the environment and any rules governs it. In the last decade many researches have been done to solve problems in path finding. There are some algorithms that stand the most outstanding, namely they are A* [3], Dijkstra [4], PRM [5], and RRT [6]. And the biologically inspired algorithms which are more popular currently are on the rise such as ANN [7], GA [8], PSO [9], and RL. In addition, RL algorithms have also received an attention thanks to its efficacy in game theory, robotics and control engineering [10].

## III. Methods and Resources

In this research, tests are done in simulation with codes written in python and achieved success. In our model, there are 14 states and 2 output actions. There are 11 million steps in average, and it took nearly 8 hours to calculate all of them. Our model is trained on GTX 1080 TI, which has a total 3,584 CUDA cores .Only 1678 Megabytes are used over 11 Gigabytes. The main reason why its size is small is that our neural network is not deep as the ones used on normal deep learning. If we were to use CPU instead of GPU, it would take days to get the same results. For our simulation, we created an environment on our computer. Our agent is supplied with the ability of interacting with the environment in order to gain information about it. All interaction is conveyed through python. Python has been chosen because of its simplicity and wide usage. We choose computer sample size huge, our intention was to have enough space for the future dogfights that will take place. Our environment was $4000\ m\ x\ 4000\ m.$ Our UAV's sachet length is 50 meters, it was given that size to make it easier to observe its maneuvers in the environment. After creating our environment and our agent, we have given information that our agent needs, they are the followings: 8 different LIDAR information of UAV, speed of UAV in x coordinate, speed of UAV in y coordinate, acceleration of AUV in x coordinate, acceleration of UAV in y coordinate, steering angle of UAV, and angle between UAV and environment. In total there are 14 information that are conveyed at a given time to our agent. Our agent decides on its movements accordingly to this information. But in the first 200 episodes, our agent just randomly moved around to make a discovery and fill its experience replay. Then it decides on the actions with the help of neural network.

### A. Methods Used in This Project

TD3 + HER are used in this project, TD3 [11] was published in 2018. The main difference of this method when compared to other methods is that even if it uses actor-critic as a base, we can't deny that classical methods like DDPG [12] have some problem. Those problems are mainly an overestimation, it basically can be explained as giving actions to agents in places it did not present, so this is the main reason why TD3 + HER come up with better results when compared to DDPG and classical actor-critic method. We tested DDPG at the beginning of the project, but we did not able to perform even basic actions in DDPG, that's why we have moved to TD3 instead.

### B. Analysis of The Resources for This Project

It is known that DDPG achieves some great performances under circumstances, but it gets halted because of hyperparameters and tunings. Q-functions sometimes overestimate Q-values, this dramatic increase causes policy breaking, since it creates a gap in errors in Q-function. On the other hand, Twin Delayed DDPG (TD3) addresses this issue by providing three critical tricks.Firstly, TD3 learns two Q-functions rather than just one, that's why it's called "twin", and takes into consideration small of the Q-values to eliminate the Bellman error loss functions. Then TD3 updates the policy and target networks less frequently than the Q-function itself. It is recommended to update once for every two Q-function updates. Then lastly, TD3 adds some noise to the specified target action, the aim of this is to make it complicated for the policy to smooth out exploits of Q-functions. Altogether, these enable TD3 to have significantly improved efficiency and performance over DDPG. We should take into consideration that; it is not possible to use TD3 on environments without continuous action spaces. It is an off-policy algorithm.

TD3 learns two Q-functions. It is known that TD3 learns two Q-functions, we will name them $Q_{\emptyset_1}$ and $Q_{\emptyset_2}$, it shows similarity to how DDPG learns a single Q-function. Now we need to work on loss function to show the difference between DDPG and TD3.

Firstly, target policy smoothing takes place. It is based on target policy to form Q-learning targets, and actions are used in this process, namely $\mu_{\theta_{target}}$. On each action, clipped noise gets added on each dimension. When the clipped noise is added, target action is clipped to lie in a valid action range (all valid actions, $\alpha$, satisfy $\alpha_{low} \leq \alpha \leq \alpha_{high}$). Then the target actions are formed like this:

$$a'(s') = clip\left(\mu_{\theta_{target}}(s') + clip(\epsilon, -c, c), a_{low}, a_{high}\right), \epsilon \sim N(0, \sigma) \quad (1)$$

Target policy smoothing acts as a tool to make algorithm regular. It solves the problem that may happen in DDPG: if Q-function approximator becomes absolute, or starts outputting incorrect data, the policy will take an action and fix the incorrect behavior. This is actually done by smoothing the Q-function values on similar actions, this is why target policy smoothing is designed.

Secondly: The smaller value will be taken from both Q-functions used on a single target. This is called clipped double Q-learning.
Its formula as follows:

$$y(r, s', d) = r + \gamma(1 - d) \min_{i=1,2} Q_{\emptyset_{i,target}}\left(s', a'(s')\right), \quad (2)$$

and both are learned by regressing to the target:

$$L(\emptyset_1, D) = \underset{(s,a,r,s',d)\sim D}{E}\left[\left(Q_{\emptyset_1}(s, a) - y(r, s', d)\right)^2\right], \quad (3)$$

$$L(\emptyset_2, D) = \underset{(s,a,r,s',d)\sim D}{E}\left[\left(Q_{\emptyset_2}(s, a) - y(r, s', d)\right)^2\right]. \quad (4)$$

.

The smaller Q-value is used for the target. This prevents overestimation in Q-functions. Lastly: $Q_{\emptyset_1}$ is maximized,

$$\underset{\theta}{\text{Max}} \underset{s\sim D}{\text{E}}\left[Q_{\emptyset_1}\left(s, \mu_\theta(s)\right)\right]. \quad (5)$$

Well you may not see any noticeable differences in theory from DDPG. But in TD3, it's Q-functions are not updated as frequently as they are in DDPG. This solves the volatility problem that occurs in DDPG which is caused by updates.

TD3 follows a deterministic approach in an off-policy way. On condition that agent try to explore on-policy, it will not try a wide variety of actions for finding useful learning signals, this is caused because the policy itself is deterministic. In order to make TD3 policies explore in a better way, noise is added to their actions at training time, or typically uncorrelated mean-zero Gaussian noise. Reducing the scale of the noise helps with getting higher-quality training data, though we did not do this in our research, the noise is kept at a fixed scale. Side note, noise was not added to the actions to see how the policy handles exploits it learnt during the test time.

HER [13] enables RL algorithms to be applied to the problems with sparse and binary rewards. Also, this can be run with arbitrary off-policy RL algorithms. In the well-known DQN algorithm, past experiences are buffered, and they are used to stabilize training. The buffer works with a principle that records past states, actions taken, and the rewards received, and the states that are observed. In order to apply this scheme to multi-goal setting, goal should be saved before the state, and machine should learn the value of state-goal-action triplets. Now this question arises, if the circumstances were different and we add fictitious data, what would happen? Well HER answers this question. HER, follows these steps, agent tries to perform an episode to reach its goal state G from its initial state S, but it cannot achieve state G and results in state S' instead. It is cached in trajectory in replay buffer as:

$$\{(S_0, G, a_0, r_0, S_1), (S_1, G, a_1, r_1, S_2), (S_2, G, a_2, r_2, S_3) \dots \dots \dots (S_n, G, a_n, r_n, S')\} \quad (6)$$

.

Letter “r” stands for subscript, and k stands for the rewards received at step k of the episode, and “a” is also subscript “k” is the action taken at step k of the episode. The main concept behind HER is acting as if S’ was our target goal from the beginning, and it assumes that agent has reached the goal, and got a positive reward. So, the following is also cached,

$$\{(S_0, S', a_0, r_0, S_1), (S_1, S', a_1, r_1, S_2), (S_2, S', a_2, r_2, S_3) \dots \dots \dots (S_n, S', a_n, r_n, S')\}. \quad (7)$$

But this trajectory is not real, it is set up, and its main motivation is human ability to learn from failed attempts. Also, even if it didn’t reach the target goal, since we act as if it reached, the reward is positive. Introducing imagined trajectories enables that our policy will always learn from positive rewards. Well in the beginning these imaginary states will be the only states that randomly initialized policy could reach, although they have no practical usage. Frankly, with the help of function approximation it will be ensured that our policy will reach the states similar to ones seen before. This is achieved by deep learning, and generalization. The agent will reach small are at first, but it will expand the reachable area gradually till it learns how to reach the goal states that we aim for.

Curriculum learning which is so common in deep learning has many similarities with the process we explained here. Curriculum learning allows agent to work on the real task, to promote learning of neural network, but it usually fails. The best way to achieve is though, is making it work on a small scaled instance of the problem and increase the difficulty slowly till it solves the target problem well. It should be noted that curriculum learning often works in practice situations, but it requires manual engineering, a nd producing easier instances is a must. But the problem is that this is not always an achievable situation, also it might be difficult and time consuming according to the situation. When it is compared to HER, HER provides a similar outcome with less tuning and designing a specific curriculum. In fact, it is possible to consider HER as a curriculum learning process implicitly, agent is always supplied with problems that are in range of its capabilities to solve it and then increasing the spectrum to harden the difficulty.

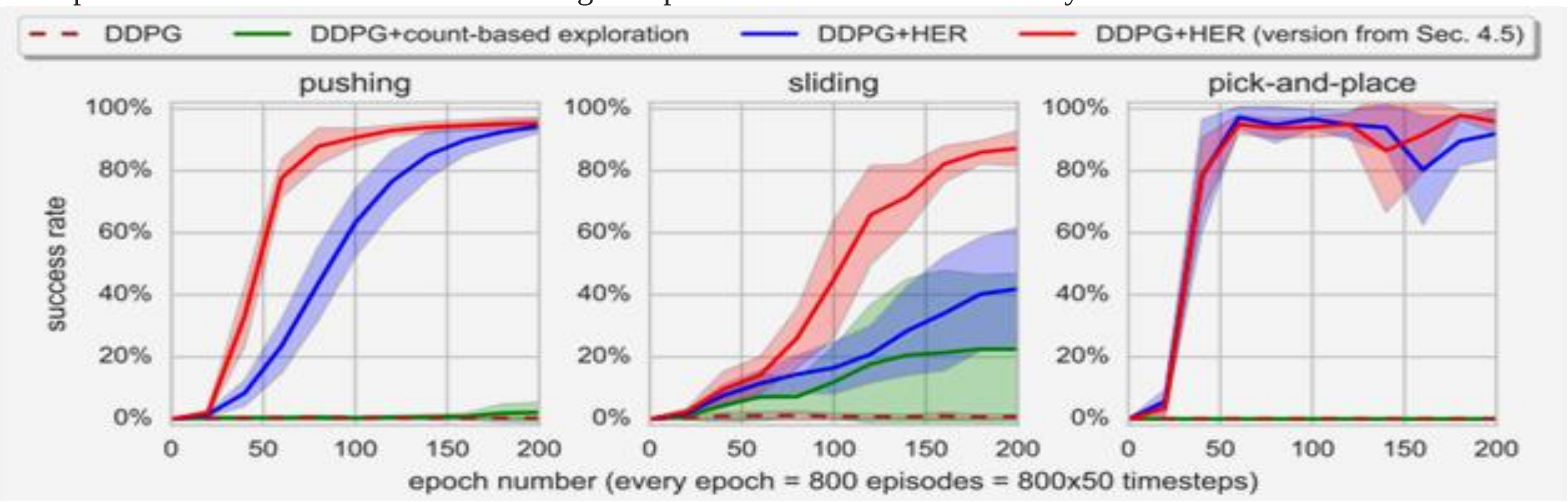

**Figure 1: Adopted from [13].**

They have tested on several occasions in which they tried to achieve several goals like picking up objects, sliding them to a certain position, these are called robotic manipulation tasks. They have given reward if the task is completed successfully, and if it failed there were no rewards. In the research they have taken DDPG as a base algorithm for HER, they claimed that HER succeeded in learning in areas other algorithms had failed (Statistics are shown in Fig.1). When a specific goal is cared about, HER provides significant performance boost over other algorithms, as long as it is provided with sufficient goals for training. This method is a superb alternative to solve important and different problems in RL applications.

## IV. Reach Target with TD3+HER

When literature research has been done, it can be seen that similar applications have been used in many different areas. To demonstrate that, there are robots that use RL in uncharted environments. In that research, it is requested from the robot to reach the goal by using q-learning to recognize the environment. Also, there is a work done on completely different sector, indoor mobile robot path planning. In this one, it is expected from the robot to reach the target in an indoor space in an optimal way [14].

### A. Overview

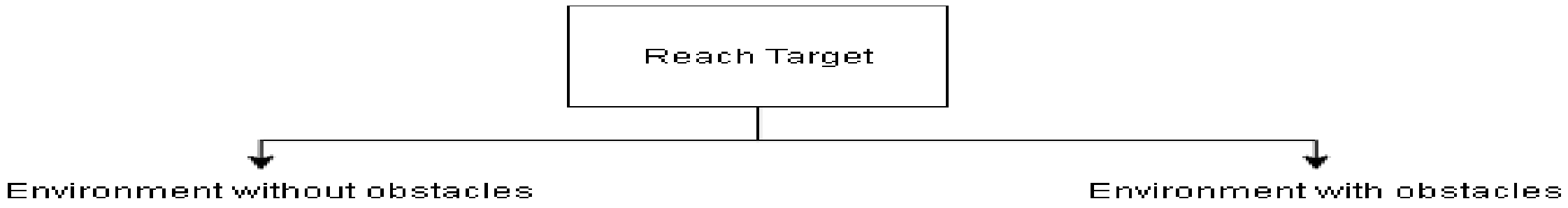

**Figure 2: In this paper we experiment in two different environment.**

In this project TD3 has been used. The reason why it has been chosen is that, it is the best suitable one for continuous actions. One property of the vehicle is that it has 8 LIDARS, these convey the total distance from the barriers that touchs laser from the point of vehicle (see Fig.3). It prevents our vehicle to crash into corners.

I defined two continuous actions, one for speed, and the other for direction. Both takes values between -1 and 1. If the speed value is higher than 0.15, it adds up the value as a velocity, but it is capped to get maximum value of 1. If the speed is between -0.15 and 0.15, it does not do any speed up or slowdown. If it is less than -0.15, it brakes and gains velocity to negative direction.

Same principle applies for the direction too. It is between -1 and 1. If the value is higher than 0.1, it turns to right, and if it's less than -0.1 it turns to left. But the turning angle is still determined by its value, to demonstrate that if it takes the value of 0.2 and it tries to turn, it steers just a little. But if it takes 1 while turning, it means that it will do an almost full turn. If the value is between -0.1 and 0.1 it will steer straight.

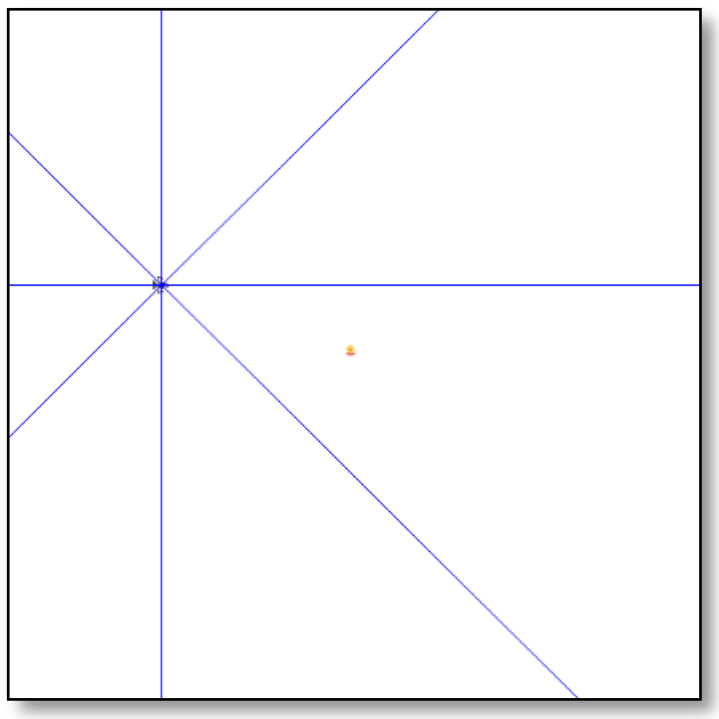

**Figure 3: This figure shows LIDAR of UAV.**

### B. Observation states

The followings are our states for this experiment: 8 different LIDAR information of UAV, speed of UAV in x coordinate, speed of UAV in y coordinate, acceleration of AUV in x coordinate, acceleration of UAV in y coordinate, steering angle of UAV, and angle between UAV and environment. Totally we have 14 states.

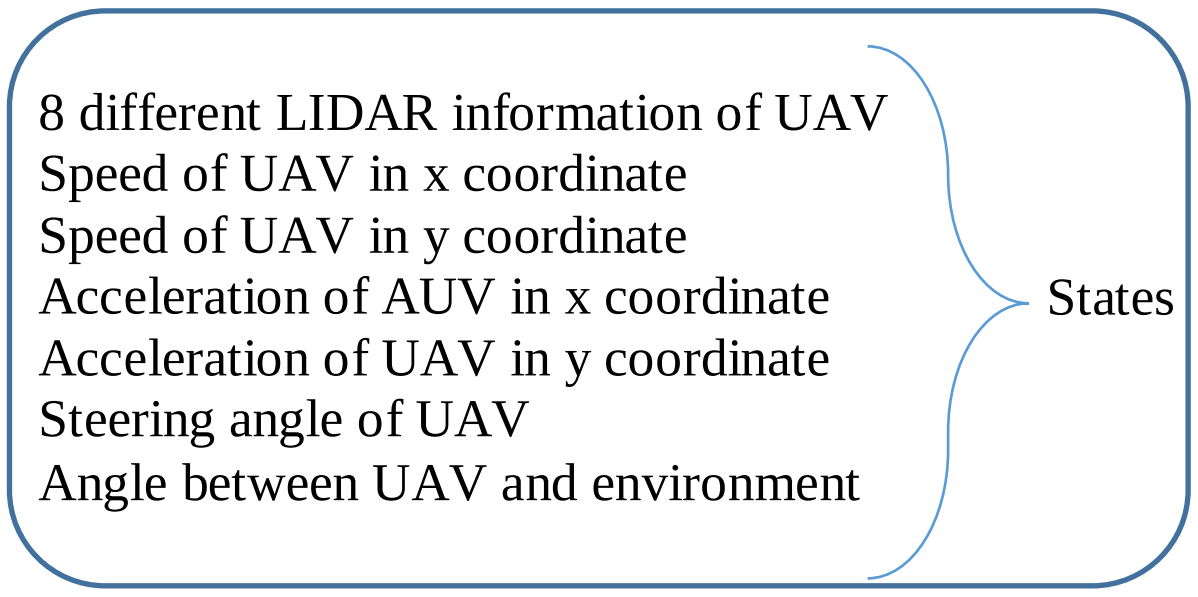

For multi agent I changed states. I reduced stated . As we can see we used only X,Y coordinates of UAV , X,Y axis , Euler Angles,Speed of UAV in X coordinate , Speed of UAV in Ycoordinate , Opponent X,Y coordinates , Steering angle of UAV , Angle between UAV and opponent.

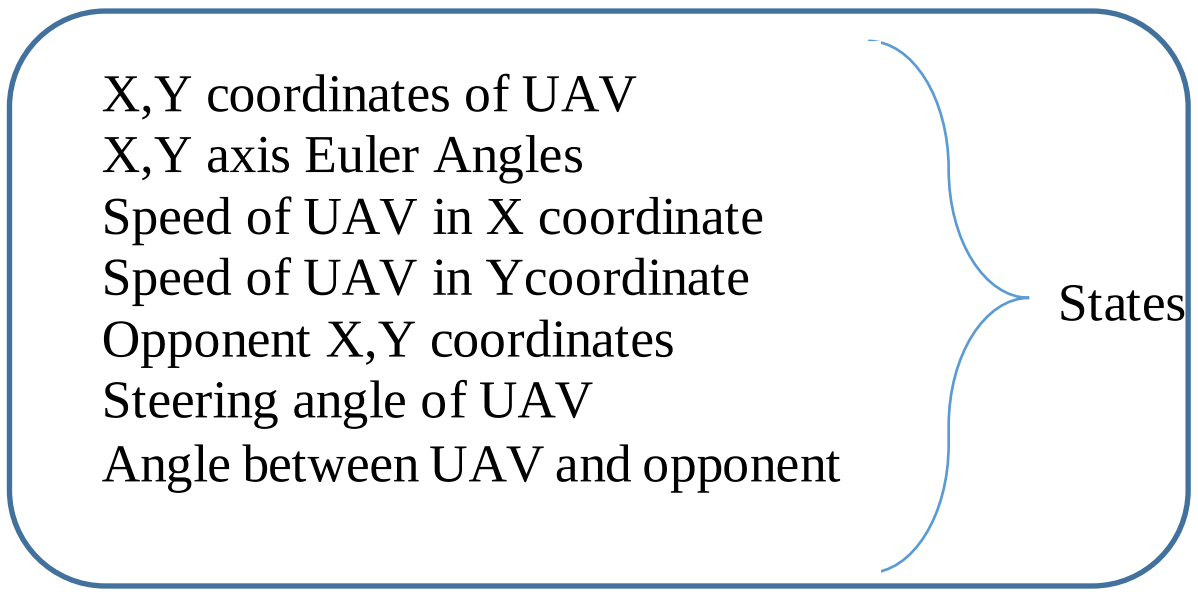

### C. Reward Function

Our environment has the size of $4000\,m\ x\ 4000\,m$ it is in a square shape with walls on its boundaries. Each time our UAV hit the wall, it got -100 points. The total rewards system works like this: each time it reaches to the goal, it gets +10.000 (ten thousand) points, but for the total time stamp in which it didn't reach the goal it received a negative reward. The formula for 2D shown as:

$$Reward = -10^{-5} * \sqrt{(vehicle_x - target_x)^2 + \left(vehicle_y - target_y\right)^2} \quad (8)$$

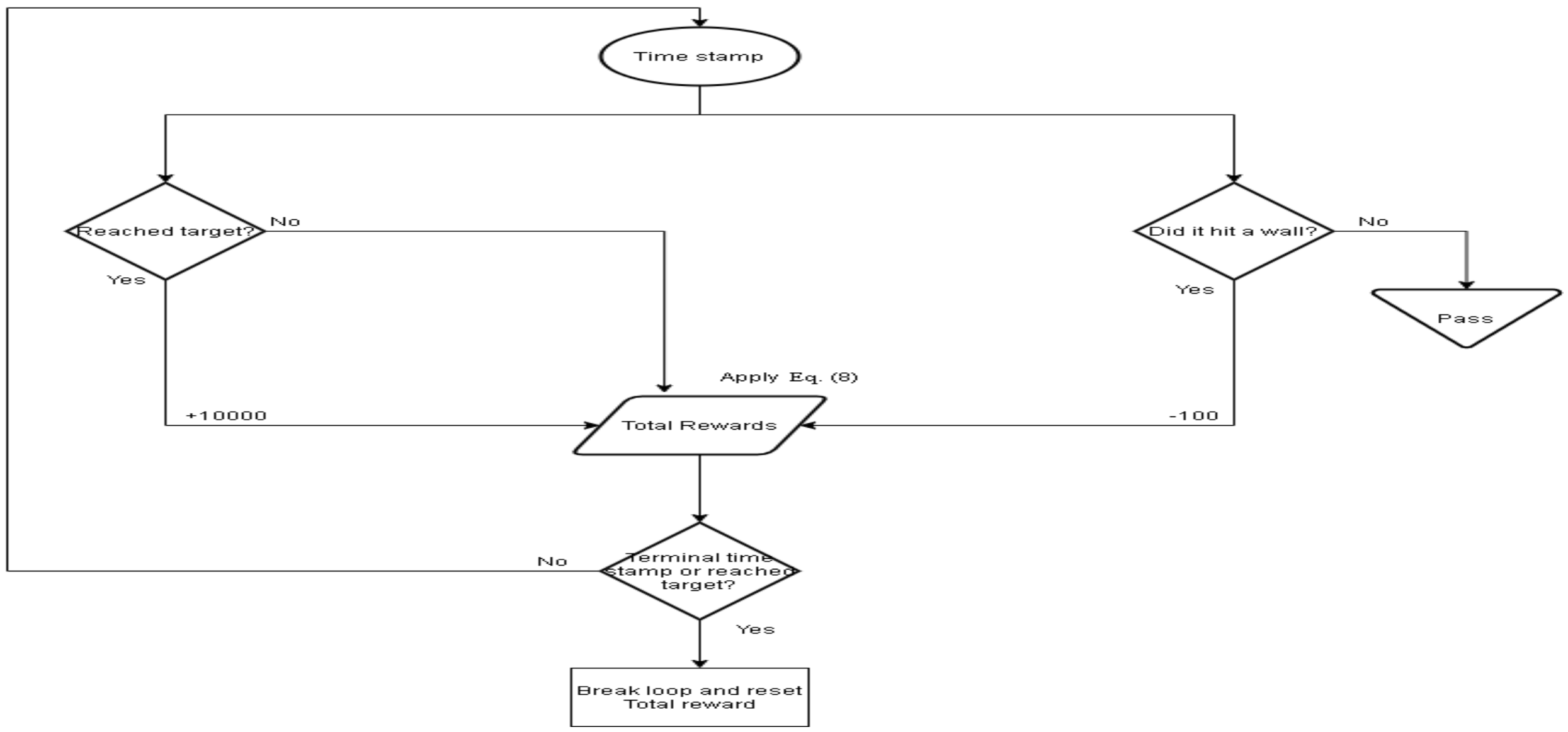

**Figure 4: Flowchart that demonstrates the reward system for each time stamp(2D).**

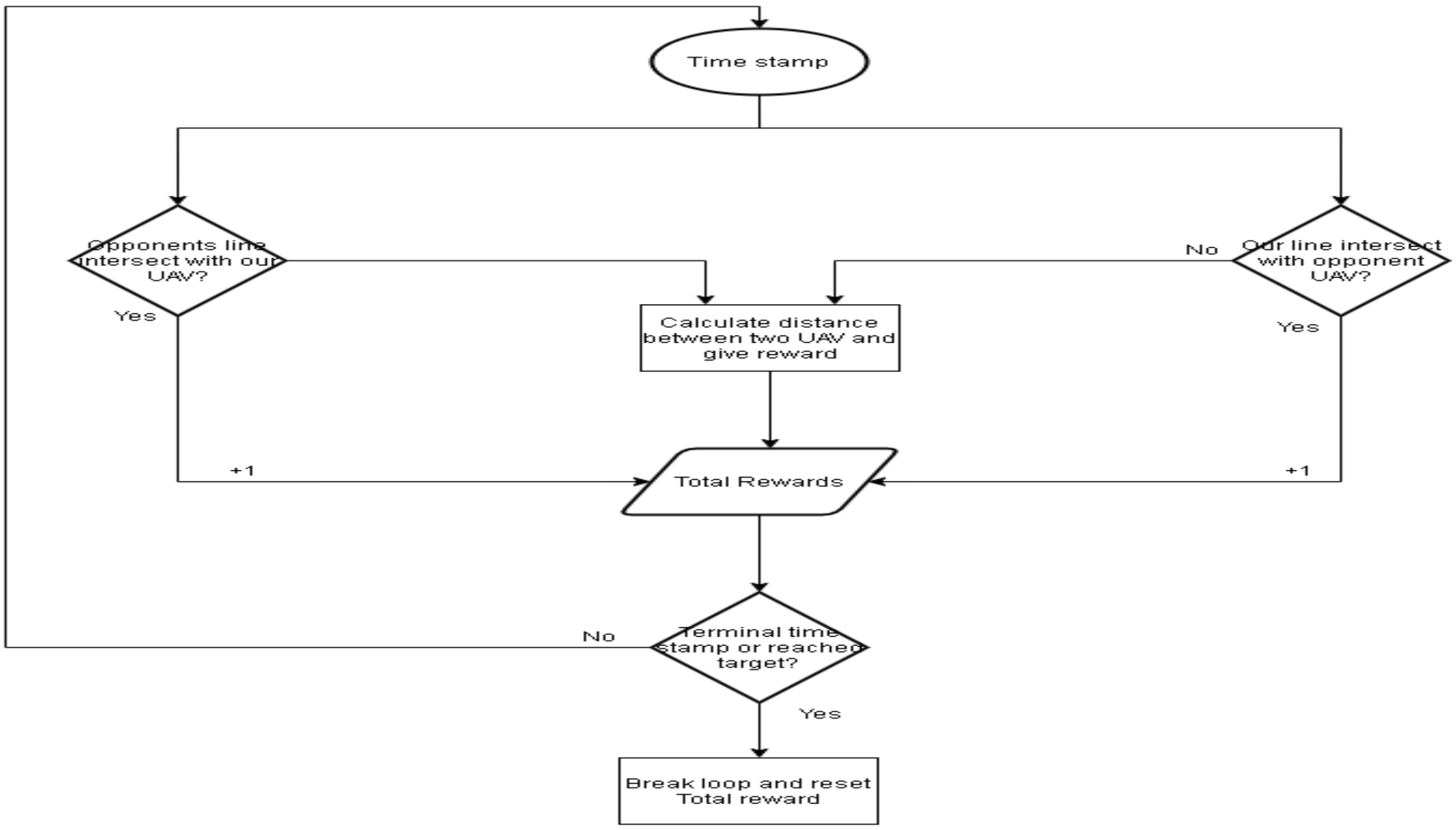

**Figure 5: Flowchart that demonstrates the reward system for each time stamp for multi agent dog-fight (2D)**.

All in all, in total there are three rewards, when it reaches the goal +10.000, if it hits the wall -100 and for each time tick it didn't reach the goal it gets negative rewards based on the distance from the goal (See Fig. 4). Rewards have been defined as this. So that it will try to reach the target as soon as possible without going too far from the goal.

For a 3D environment, reward function changed. Reward function depends on AA and ATA.

- Air Dog Fight

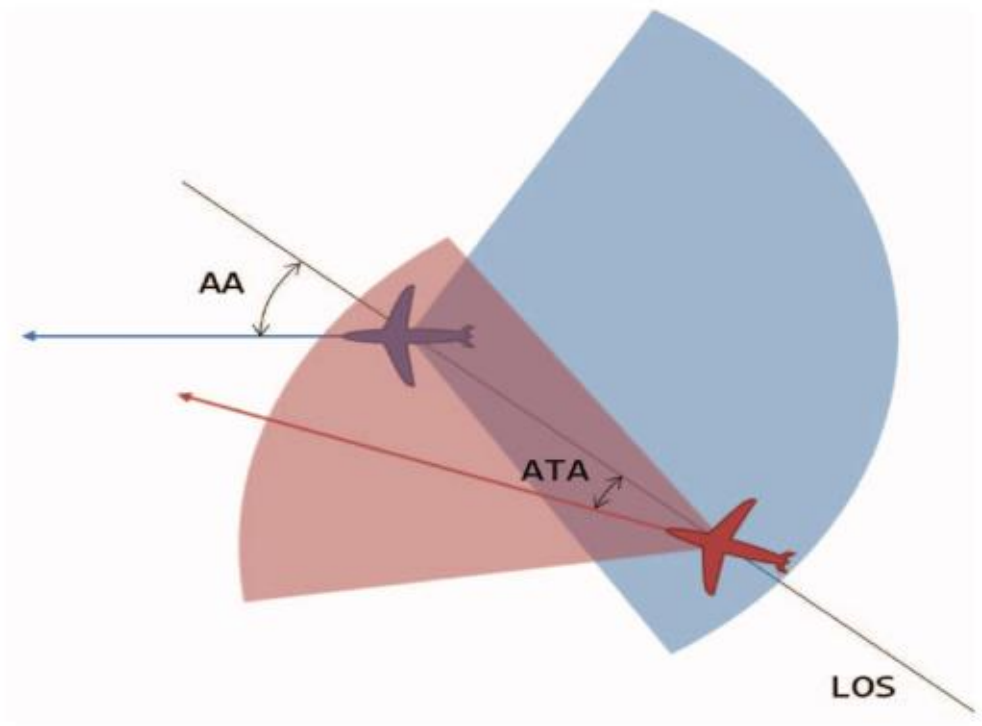

E.g. reward

**if** $R < 0.1m$ **then**
  $g = -10$
**else if** $0.1m < R < 2m$ **then**
  **if** $|AA| < 60°$ and $|ATA| < 30°$ **then**
    $g = 1$
  **else if** $|ATA| > 120°$ and $|AA| > 150°$ **then**
    $g = -1$
  **end if**
**else**
  $g = 0$
**end if**

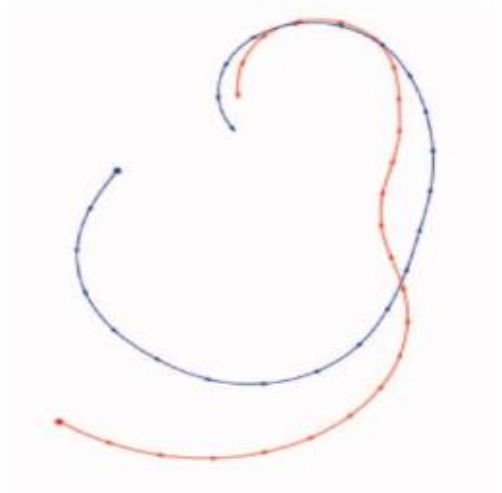

**Figure 6: Reward function for 3D Environment.**

### D. Model

Model is firstly tested in an environment in which all sides are covered with walls, it achieved intended goals. 200 episodes has been used for exploration and 30 was used for validation. At the end of every 100 episodes, 30 episodes will be taken into consideration to evaluate the model's performance(validation) . Each episode consists of 1000 steps. Model cannot learn good if exploration has not been done before, thus exploration is vital and its significance is cannot be deniable. In exploration phase, our agent roams around randomly and fills its buffer. After the exploration training phase takes place. With the help of the buffer that was filled in the exploration phase, model checks the buffer and completes the training with the help of it. In the researches buffer size is 1 million, so this research's buffer size is also 1 million.

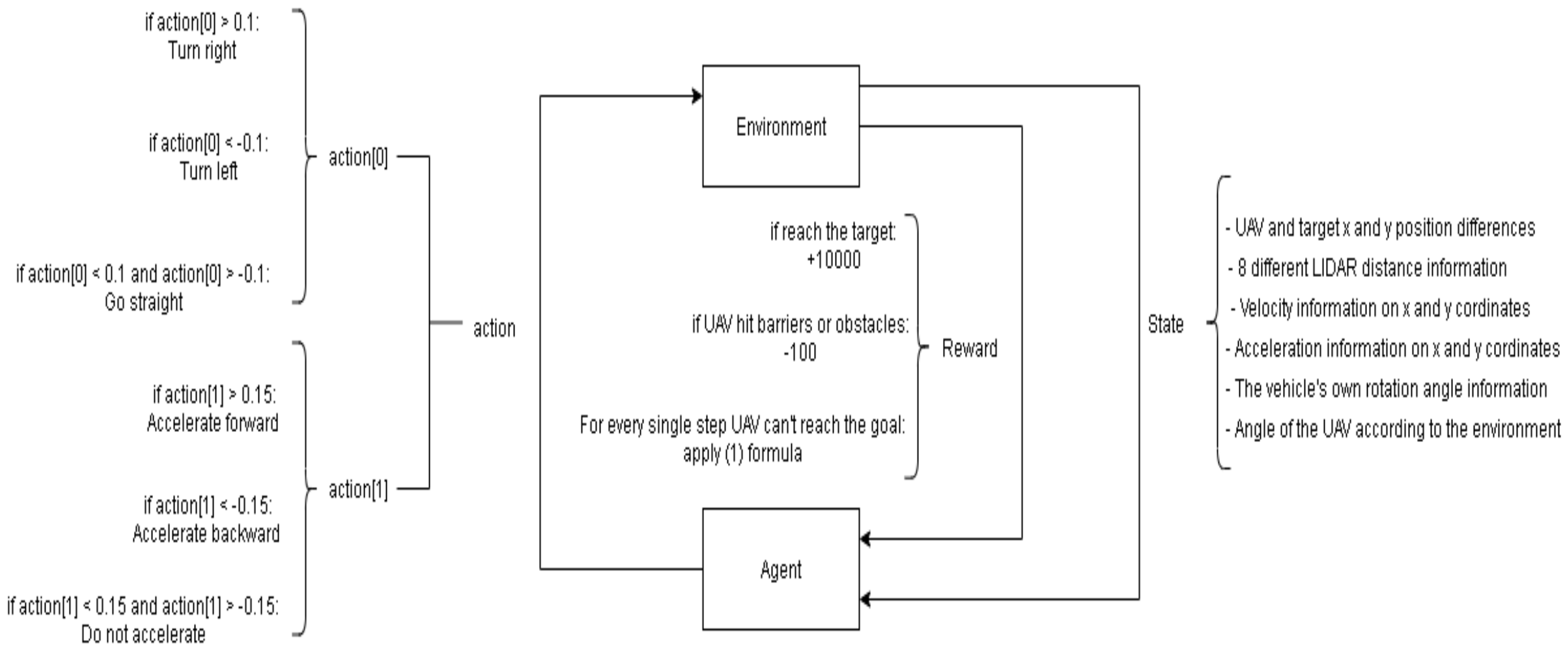

**Figure 7: Overwiev of the model (2D).**

**E. 3D Environment and features**

In order to transfer the model from 2D dimention to 3D dimension, Python Game Library has been accompanied, though due to the fact that outcome of the pygame conversation was not satisfactory, HARFANG3D platform has been used instead. HARFANG is a software framework in which you can manage, display complex 3D scenes, play sound and music, acces VR devices like Oculus Rift, it is specifically designed for modern multimedia application development. HARFANG is a platform which has the fundamental aspects of flight dynamics. The first step was establishing a link between our agent and the platform, we have achieved this by using a network client which has been created by our team, the distinguishing factor of the network client is that it is completely modifiable by the requisities of our needs.

There are two diference variences in the network. In the former one, an agent can work on different environments simultaneously and it can collect data from each environment in each time step to update itself, thus sending different actions to each environment. In the second varience, multiple agents can work on seperate environments in which they have their own seperate observations, and learning situations. In both variances, they have advantages over one another. In the first one the agents can observe different cases which makes the training come out to be better because the transitions that are sent to the buffer are distinct from one and onther thus leading a more variant observation. Whereas in the second variant, it is possible to train more than one agent. Actually, this can be beneficial for giving an input of different states then deciding on which agent works better on the steps. In this way it is possible to find the agents whom work best on particular states and actions, then making them work on the variance 1 to train them efficiently.

When the aeroplanes that are used in the HARFAN environment are inspected, it is possible to see that there is a wide range of aeroplanes to choose from. The famous ones are F-16, TF-X, Euro-Fighter and Rafale which are widely-known and recognised aeroplanes. Though in the background, the dynamics and SASs are based on that of f-16's. Here areTF-X and F-16 that have been used in this project.

.

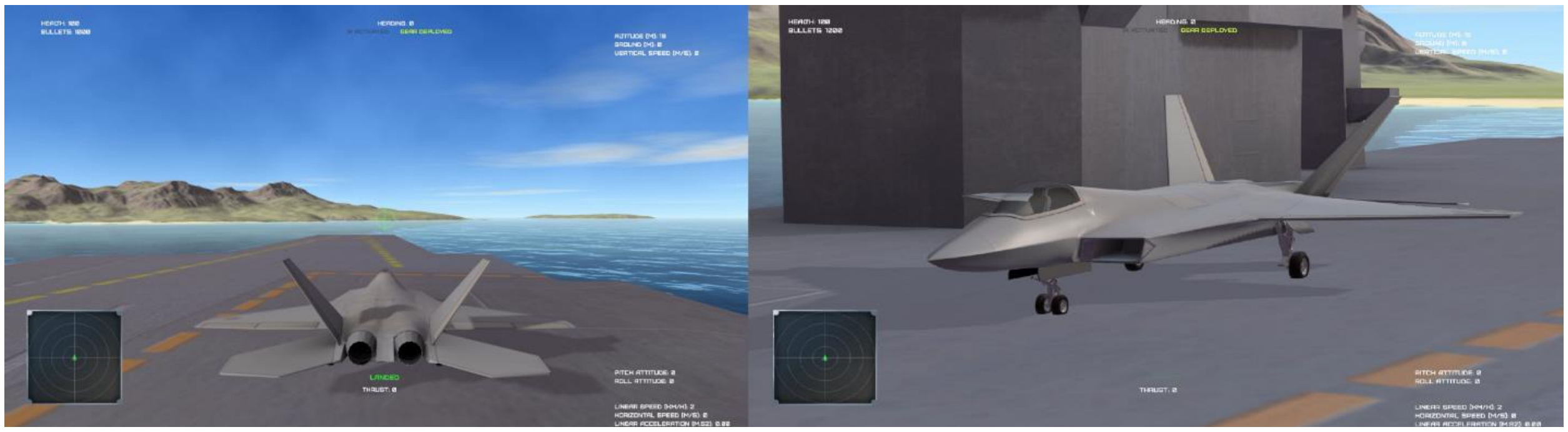

**Figure 8: Turkish Fighter TF-X.**

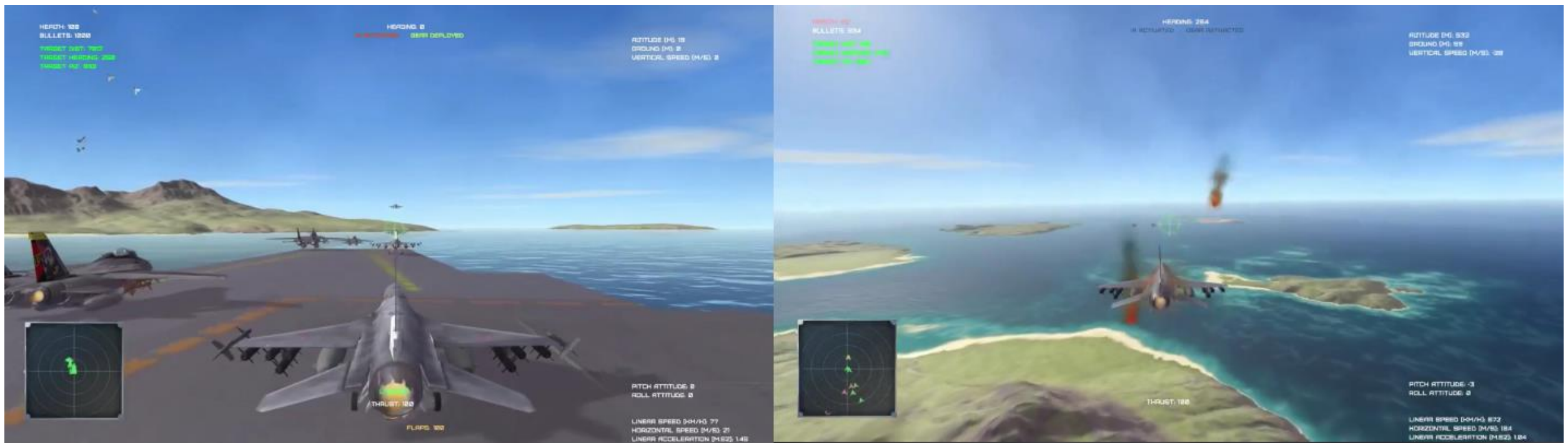

**Figure 9: F-16.**

HARFANG simulation game supports dog-fight scenes with different number of planes raging from 1v1 to 12v12. This enables the development of a multi-agent, grouped, competitive model. The main objective of this scenario is to keep the number of allied aero plane higher whereas keeping the number of opposite team's aero plane lower than their's.

Primarily it has been tested on a 1v1 scenario and actions were given. Nine states have been given to the agents, aero plane's and its own 3 axis in Euler angles , opposing aero craft's and its own horizontal and vertical speeds, its ammunition information, opposing plane's and its own coordinates in X,Y,Z axis. For the actions, there have been 5, acceleration, brake, elevator, rudder and aileron. The same states and actions were given to the opposing aeroplane, but it did not work out. Then imitation learning has been used to achieve the opposing plane to learn how to fly.

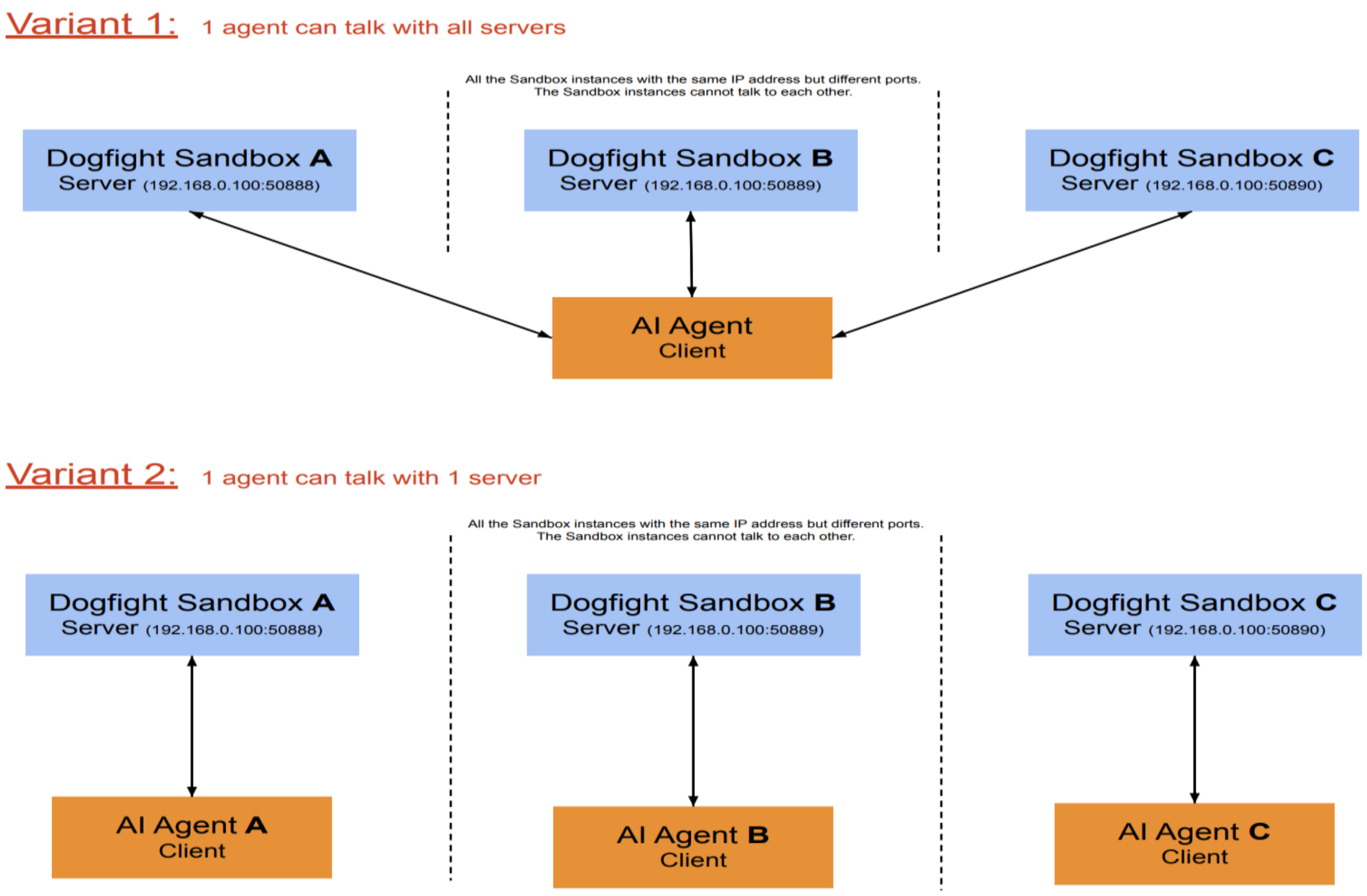

**Figure 10: Variations of training.**

### F. Behavioural Cloning

The simplest form of imitation learning is behaviour cloning (BC), which focuses on learning the expert's policy using supervised learning. An important example of behaviour cloning is ALVINN, a vehicle equipped with sensors, which learned to map the sensor inputs into steering angles and drive autonomously. This project was carried out in 1989 by Dean Pomerleau, and it was also the first application of imitation learning in general.

The way behavioural cloning works is quite simple. Given the expert's demonstrations, we divide these into state-action pairs, we treat these pairs as i.i.d. examples and finally, we apply supervised learning. The loss function can depend on the application. Therefore, the algorithm is the following:

$$1.\, Collect\ demostrations\ (\tau^*\ trajectories) from\ expert$$

$$2.\, Treat\ the\ demonstrations\ as\ i.i.d\ state - action\ pairs: (s_0^*, a_0^*), (s_1^*, a_0^*), \ldots$$

$$3.\, Learn\ \pi_\theta\ policy\ using\ supervised\ learning\ by\ minimizing\ the\ loss\ function\ \ L(a^*, \pi_\theta(s))$$

In some applications, behavioural cloning can work excellently. For the majority of the cases, though, behavioural cloning can be quite problematic. The main reason for this is the i.i.d. assumption: while supervised learning assumes that the state-action pairs are distributed i.i.d., in MDP an action in a given state induces the next state, which breaks the previous assumption. This also means, that errors made in different states add up, therefore a mistake made by the agent can easily put it into a state that the expert has never visited and the agent has never trained on. In such states, the behaviour is undefined and this can lead to catastrophic failures.

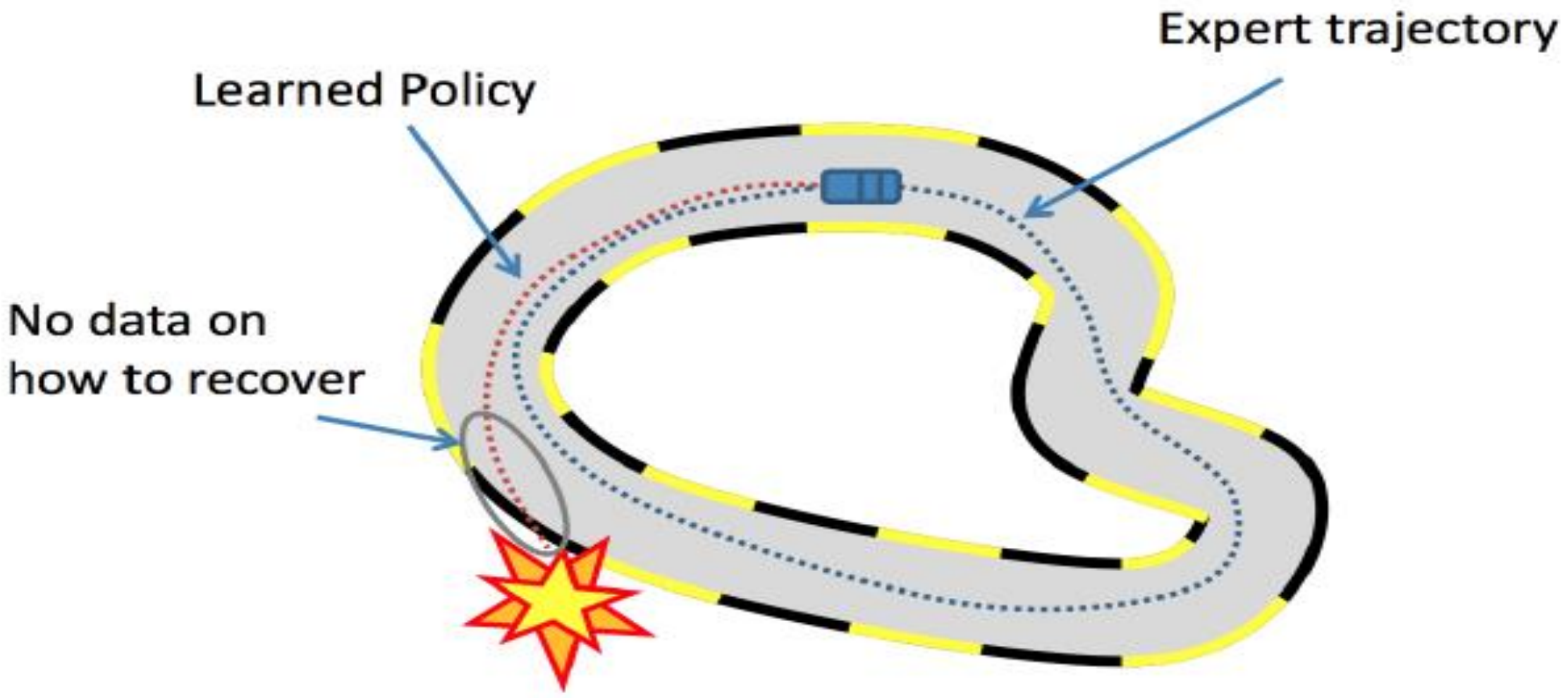

**Figure 11: Behavioural cloning can fail if the agent makes a mistake.**

Still, behavioural cloning can work quite well in certain applications. Its main advantages are its simplicity and efficiency. Suitable applications can be those, where we don't need long-term planning, the expert's trajectories can cover the state space, and where committing an error doesn't lead to fatal consequences. However, we should avoid using BC when any of these characteristics are true.

At the start of the game we will have two planes, one of which will have an NN based autopilot whereas the second one will not be based on NN. We achieved imitation learning by using autopilot which already presents in the game. If we divide the game into three sections, the first one is exploration in which our buffer gets fed with imitation learning. IN the next section two aircrafts, one of which is based on NN and the other is not, have been trained. IN the last step which is validation, we tested the aircrafts. That is how the aircrafts learnt how to fly.

## G. UAV

Our UAV starts at a random point in the environment, when it needs to turn, it must have some length itself to be able to turn in a circular angle. In mathematics the branch that deals with motion of objects is called "kinematics". It only takes consideration in velocity, acceleration and position in the space. We can say that a physical body is just a moving pixels or points on the computer screen. So that components that are in full dynamic model like engine, gears, transmission, tire friction and will not be taken into consideration. Our UAV has dubin vehicle dynamic properties, in order to move back and forth, values between -1 and 1 will be given. According to this value, it will accelerate. Acceleration formula is defined as:

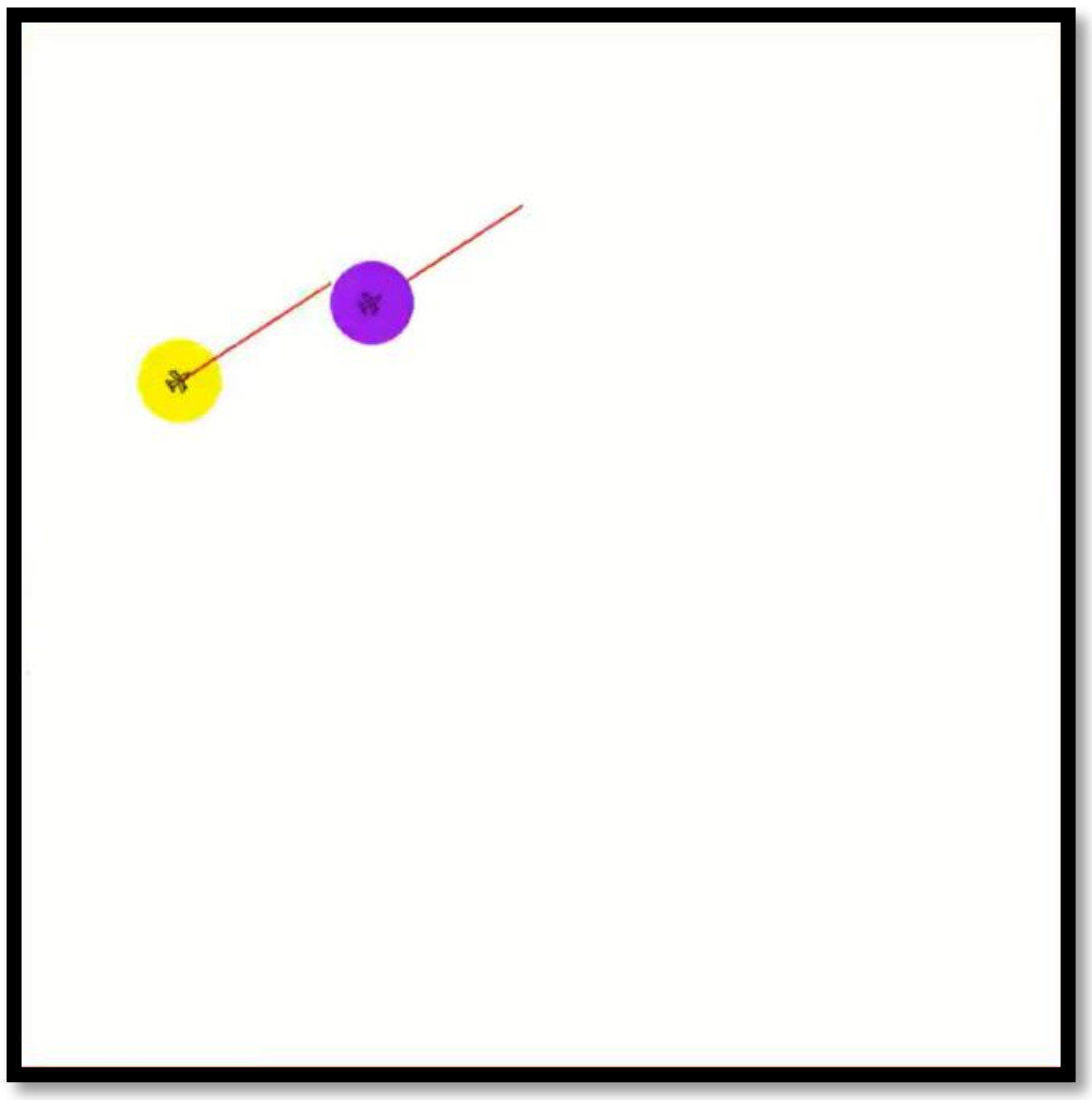

**Figure 12: Environment with Obstacles .**

$$acceleration{+}{=}\ 500 * action. \qquad (9)$$

If the action value is 1, it will fully accelerate, to illustrate that if the action is 0.2 its acceleration will be

$$500 * 0.2 = 100\ \mathrm{m/s^2}. \qquad (10)$$

If the action value is between -0.15 and 015, it will not accelerate. But if the value is between -1 and -0.15 it will accelerate backwards, thus having a negative velocity. Also, values between 0.15 and 1 means positive acceleration. The next action is turning left, right or going straight. The second action takes values between -1 and 1. This time if the value is bigger than 0 it will turn right, if it is lower than 0 it will turn left. Again, it will turn accordingly to the value it takes during turning. If the value is 1, it will make 60 degrees turn, which is the maximum value. So, if it's 0.5, it will turn 60*0.5 degrees, which means 30 degrees. Straight action is not defined because it is negligible, let's say the action value is 0.01, it needs to turn $60 * 0.01$ which is equal to 0.6 degrees, and this is not significant.

Our vehicle's default speed and velocity are 0, maximum speed is 300 $m/s$, and maximum angle that it can turn is 60 degrees, maximum acceleration is 600 $m/s^2$ and sachet length is 50 meters. The reason why these values are high is that our environment is $4000\,m\ x\ 4000\,m$. Sachet length should be long because we wanted to see it's turn in circles. If there are not any acceleration (which actions between -0,15 and 0.15) is given to the vehicle, $50\ \mathrm{m/s^2}$ acceleration will be applied to the opposing direction that is currently facing. Thus, it will simulate real life motion .We can think this situation as frictional force applying to our UAV and slowing down UAV. The general scheme can be seen on the Fig.5. Our states, actions and rewards are defined.

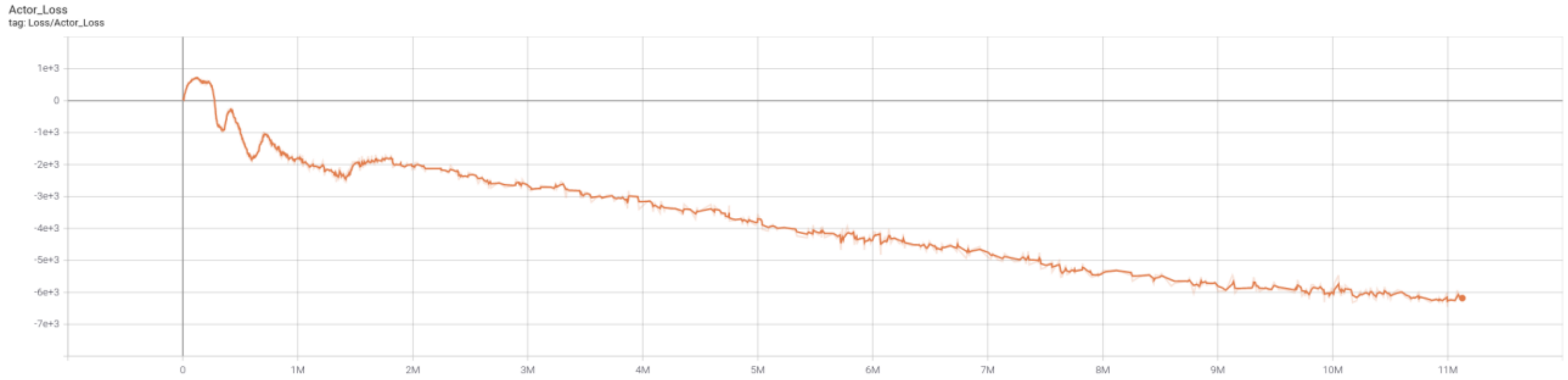

a) Actor loss

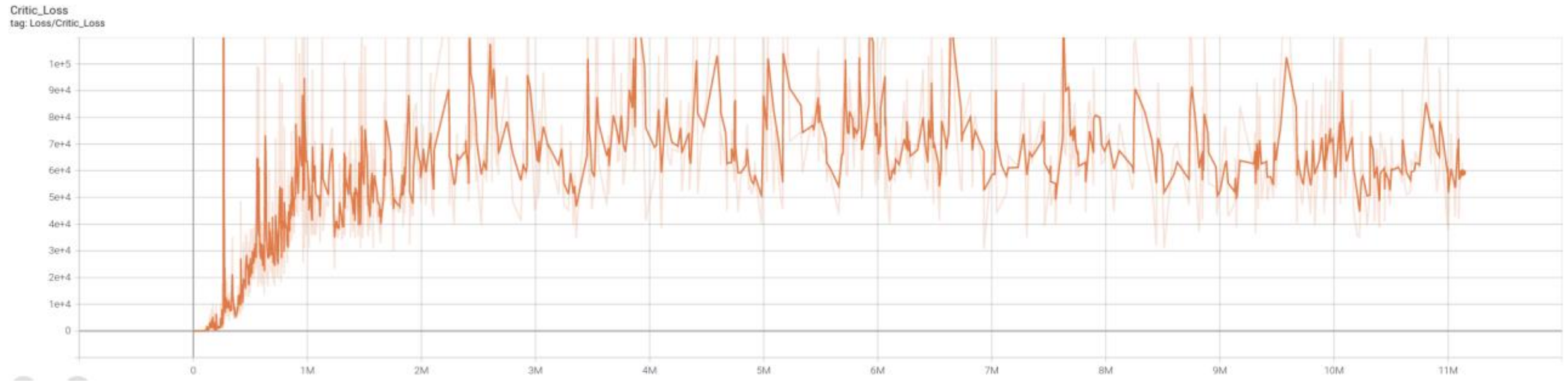

b) Critic loss

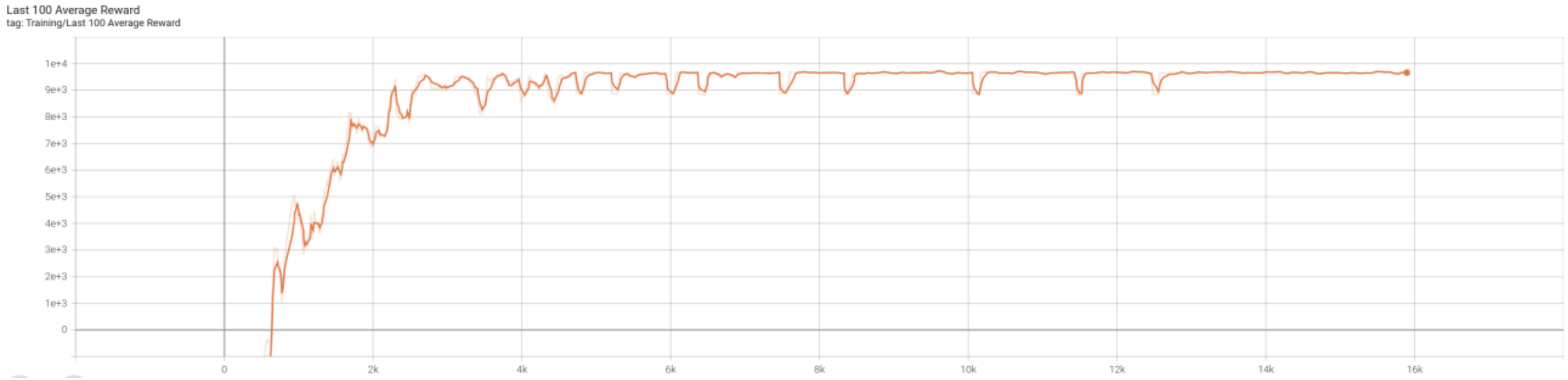

c) Average reward for the last 100 episodes

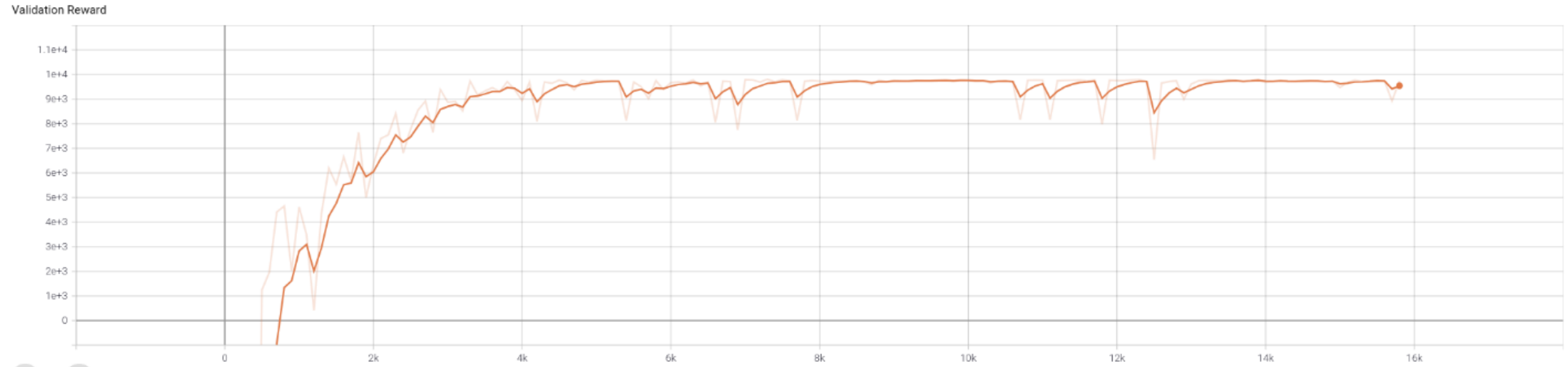

d) Validation reward

**Figure 13: Experiment results on without obstacles.**

At first, there were no obstacles in our environment and our states were: 8 different LIDAR information of UAV, speed of UAV in x coordinate, speed of UAV in y coordinate, acceleration of AUV in x coordinate, acceleration of UAV in y coordinate, steering angle of UAV, and angle between UAV and environment. We gave 200 episodes, totally 20.000 steps, for our agent to identify and cache the environment by performing random actions. In these 200 states, it didn't give any decisions they were all random.

Minimum actor loss is achieved on about 11 millionth step (See Fig. 13a). We want actor loss to be minimum as possible, the less it is the better it is. In the second figure we can see critic loss (See Fig. 13b). Critic loss has wavy behaviour, this is absolutely normal because we expect critic loss to increase gradually as actor loss decreases. In the third graph, averages of 100 episodes' rewards are shown (See Fig. 13c). And our validation reward is on 9700 which means our UAV successfully reached the goal/target (See Fig. 13d). As it can be seen that it is more stable when compared to the other graphs. In the last graph, our validation scores are shown. This is the most substantial graph of all. We can see how our agent will perform in real life situations by looking at the data shown in this graph. In its first states, our figure showed some negative outcomes, but as the progression went on its outcomes have become tremendous.

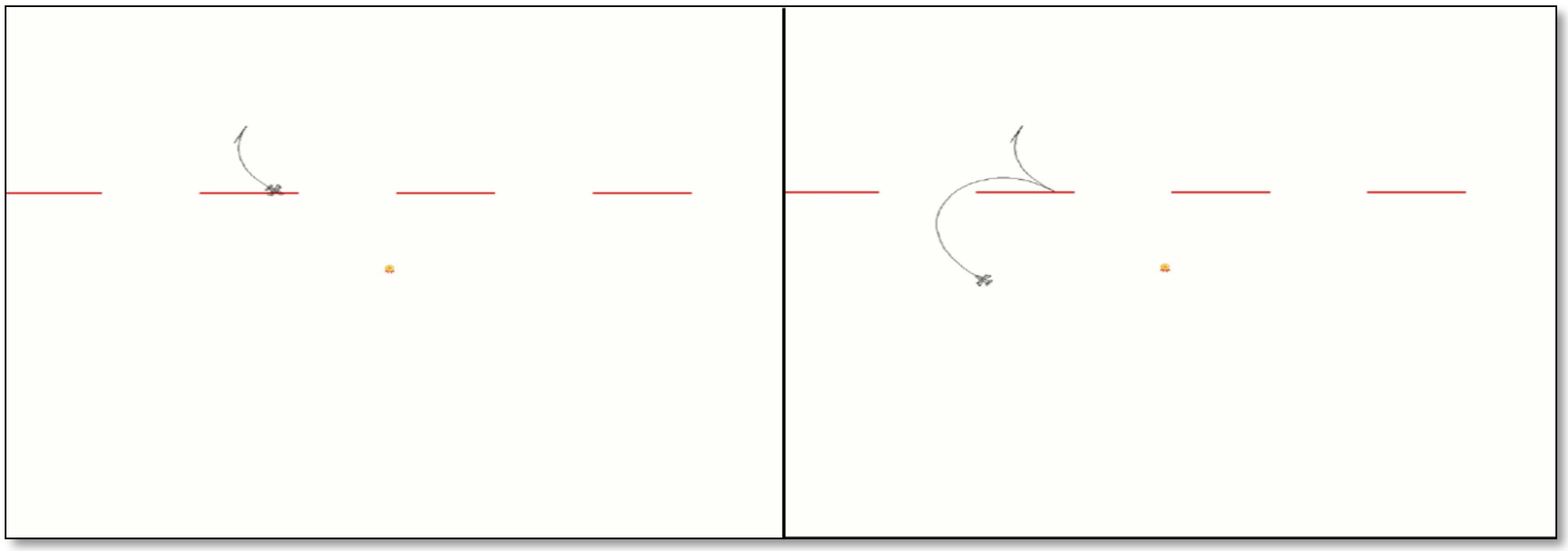

**Figure 14: Environment with Obstacles .**

Fig. 14 demonstrates two side by side progression. When our agent first hit the wall, since it can't go through it, it maneuvers, goes backwards and then tries again. With this method it is aimed to not lose any time when faced with obstacles.

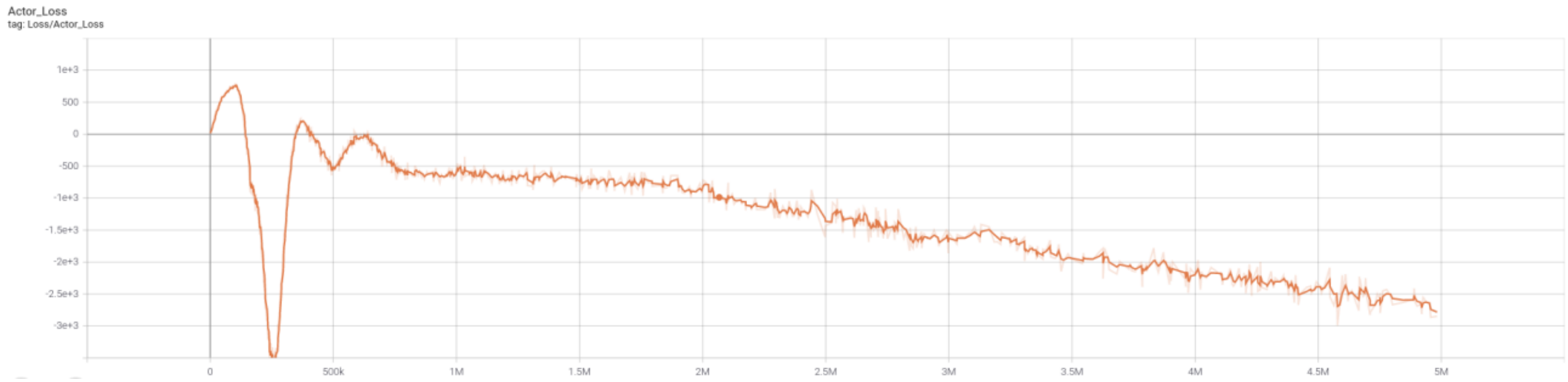

a) Actor loss

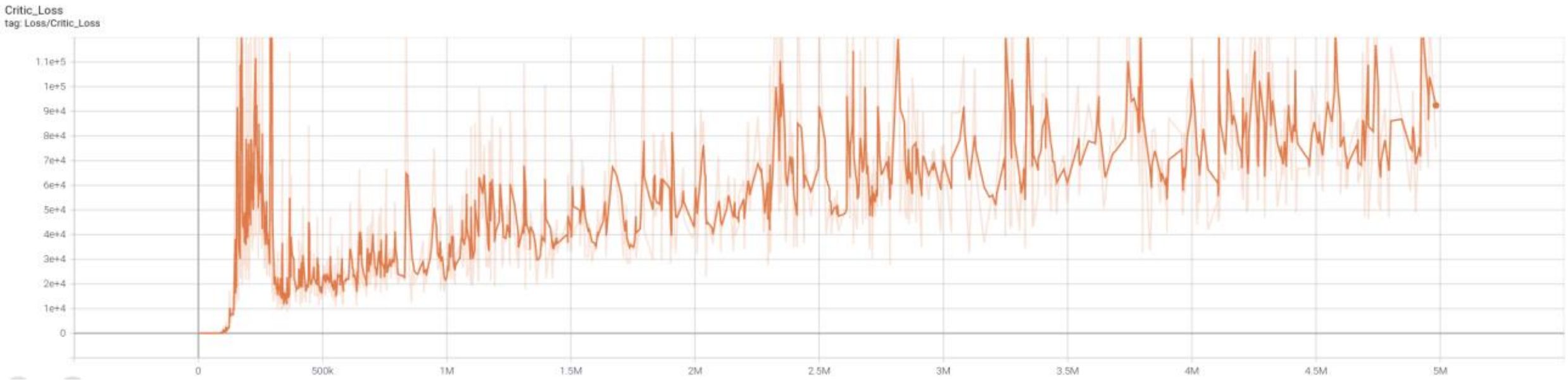

b) Critic loss

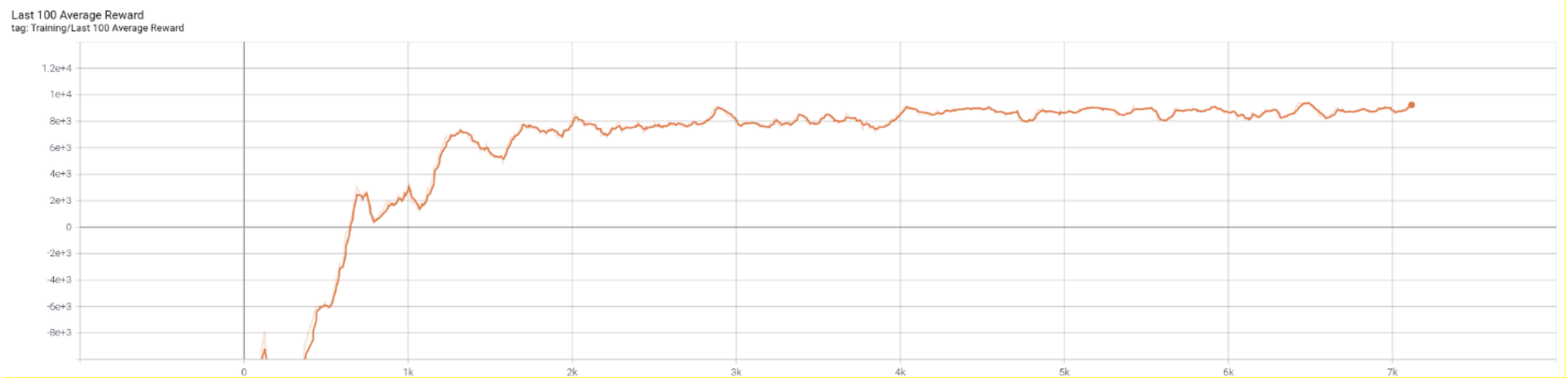

c) Average reward for the las 100 episodes

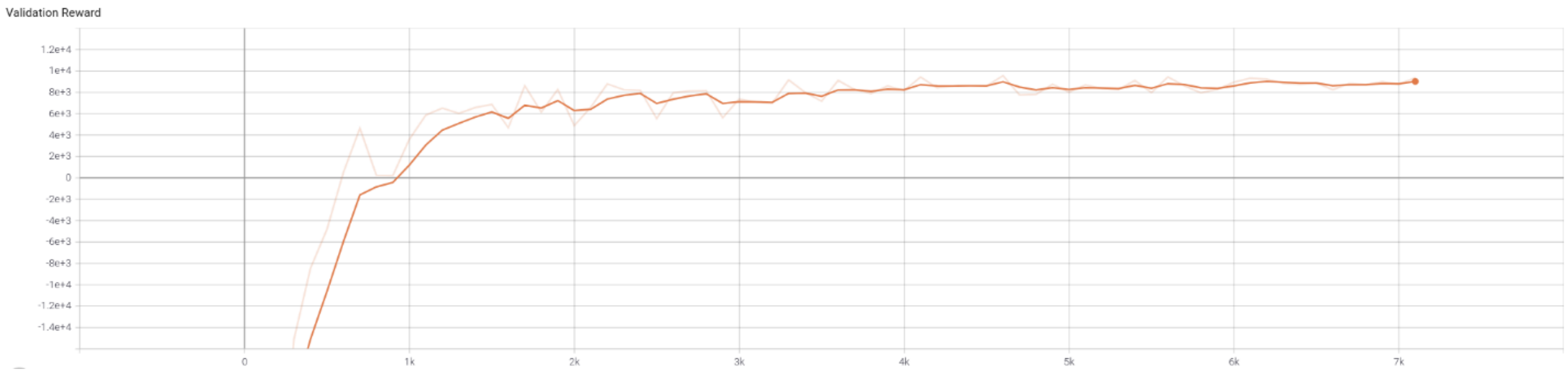

d) Validation Reward

**Figure 15: Experiment results environment with obstacles.**

For two dimensional dog-fight we have two different agent. Two diferent agent will use same buffer for getting batchs. If one agent starting to get rewards, another will loss rewards.

From figures below we can see that actor and critic loss for both agents. Also we can see average rewards for the last 100 episodes. Figures from below a,b and c figures for firt agent and d, e and f figures for second agent.

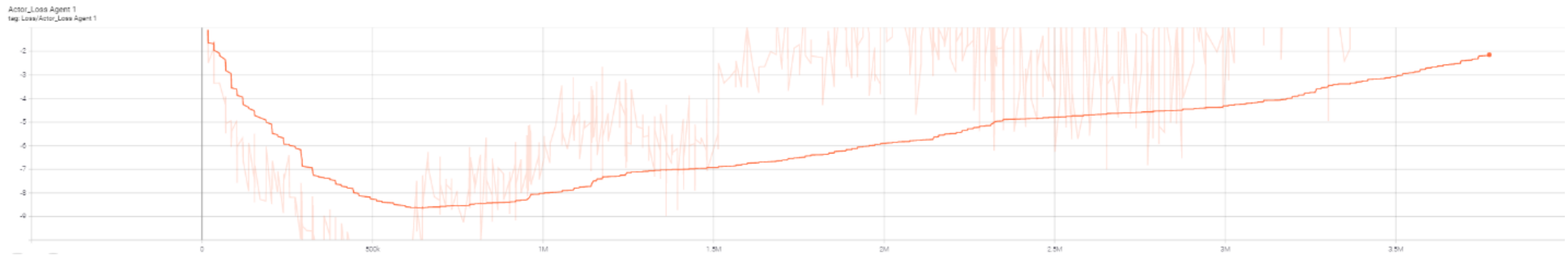

a) Actor loss

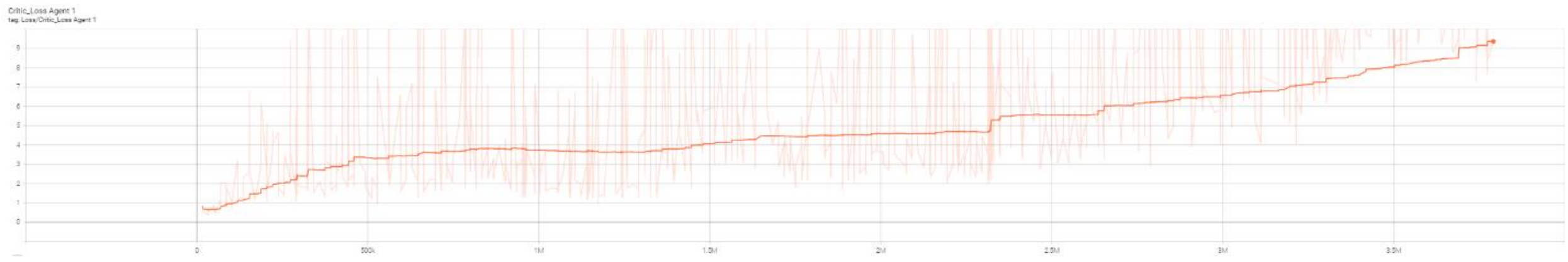

b) Critic loss

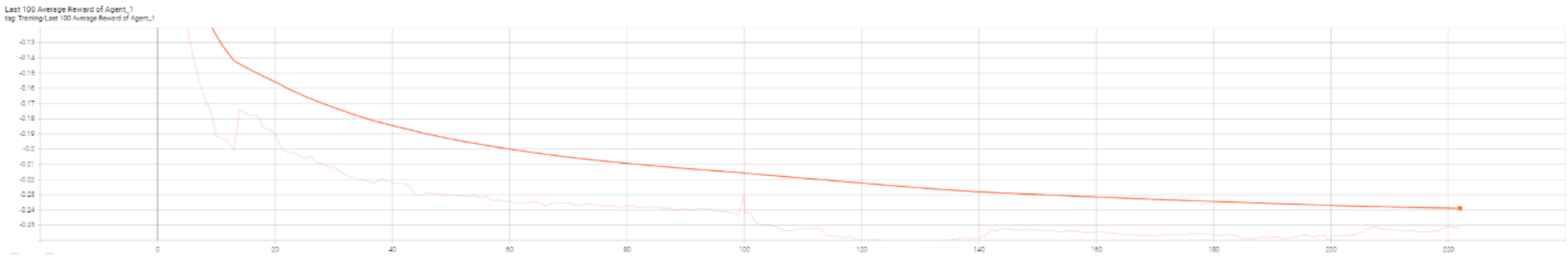

c) Average reward for the las 100 episodes

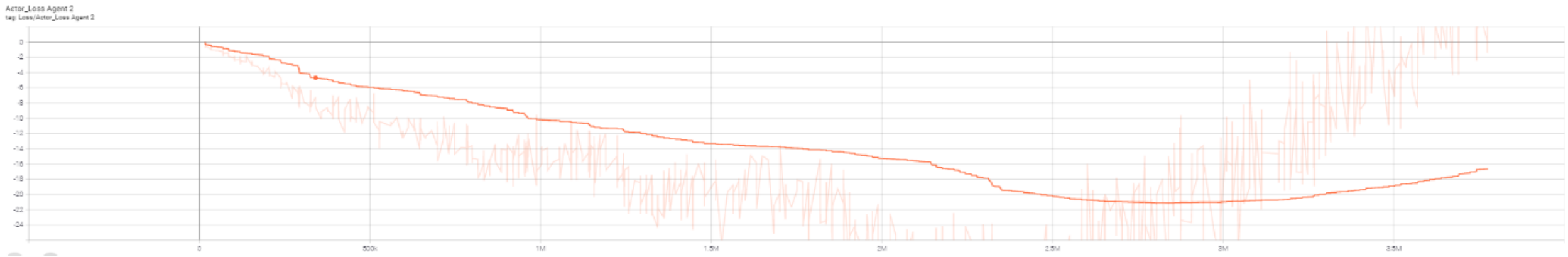

d) Actor Loss

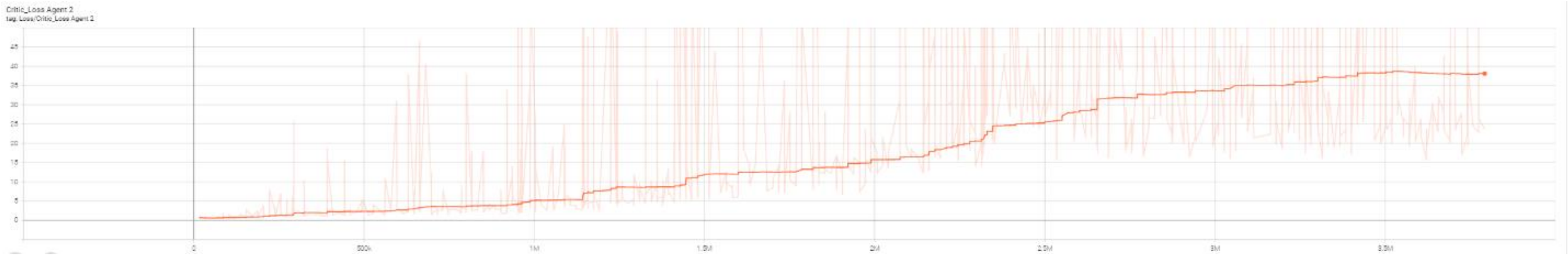

e) Critic loss

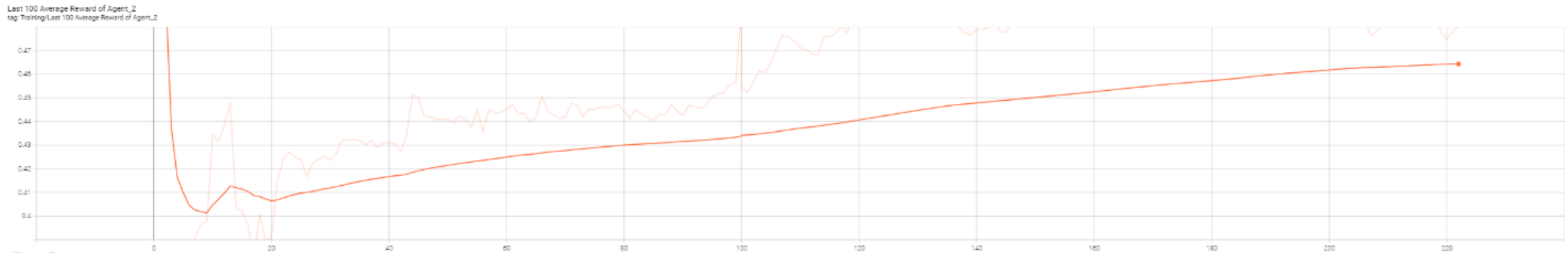

f) Average reward for the last 100 episodes

**Figure 16: Experiment results 2D dog-fight.**

When obstacles are put at the beginning of the states, it still gives a positive outcome. By using LIDARS it does not collide with the wall and successfully reaches the target. Even if our actor loss was high at the beginning, there is a gradual decrease (see Fig. 15a).Besides our actor loss' decrease, our critic increased simultaneously(see Fig. 15b). This is a good sign. When last 100 episodes are taken into account, our rewards are outstanding. We continuously got positive rewards at the end of average 1000 episodes (see Fig. 15c).But in latter episodes we also achieved some unwished results. Because every 1 action out of 10 is randomly assigned. The reason why it is like that is that we always assumed there is a better route for it to take, thus enabling it to make exploration further. When we look into last 100 episodes' reward graph, it can be seen that model learnt how to reach the target, and its proof is the line is almost linear. Also, this can be backed up with validation reward data, which revolves around +9.700 points(see Fig. 8d).

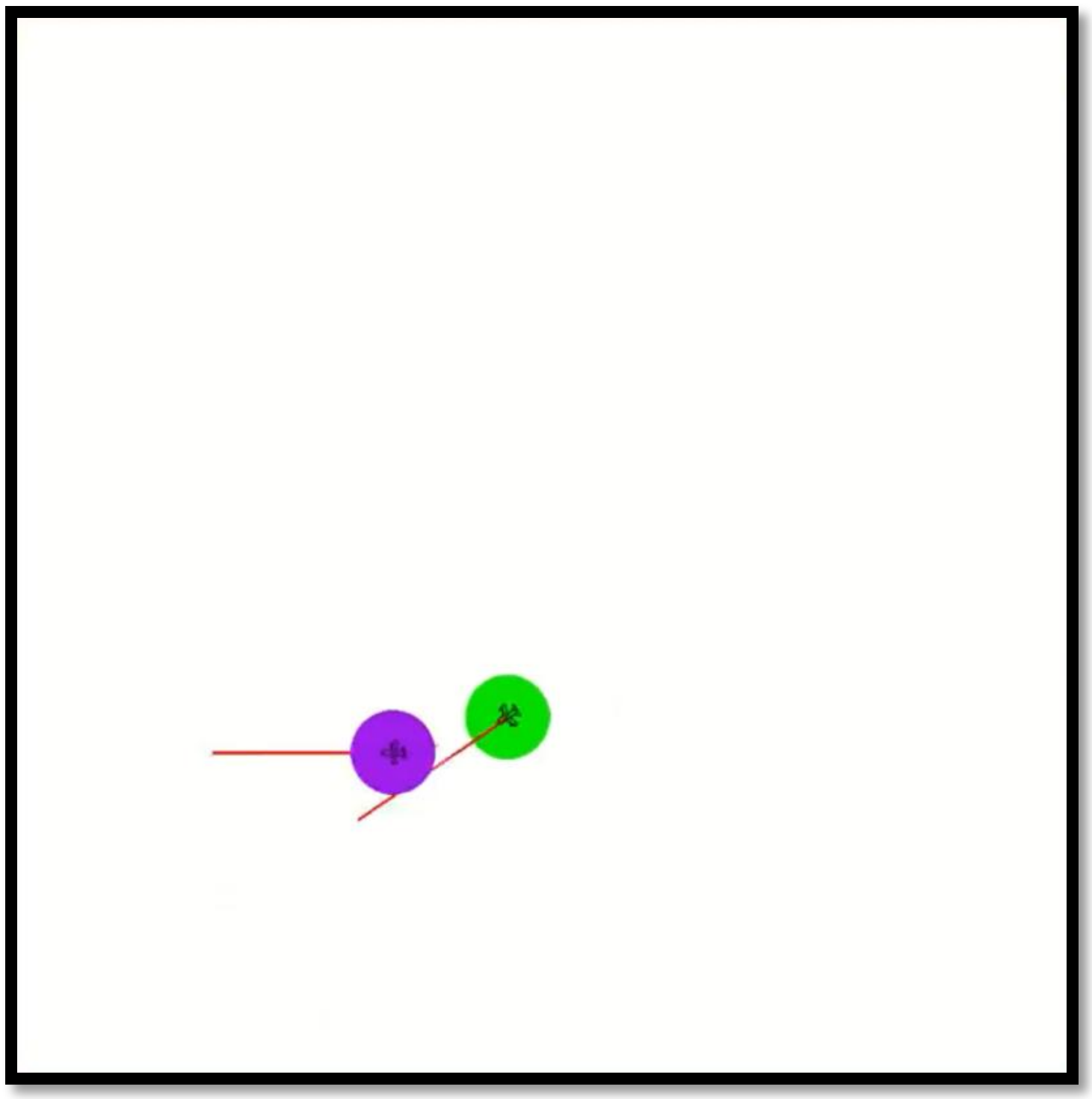

**Figure 17: Two dimensional dog-fight.**

## V. Results and Conclusion

All in all, in this research in which we have used TD3 + HER, our agent reaches its target in an optimal path. It has been given sufficient information to reach the target. In total there were 14 states information. And the actions were going forwards, backwards, turning left, turning right, and our agent successfully achieved to perform these actions. When the location of vehicle is not given to it, it succeeds when we look at the validation rewards. But if the location of the vehicle is given to the vehicle, it enables it to memorize the environment, in which hampers the ability of being successful in another environments. In the end we can say that our model succeeded, and it will show success in other environments too.

This research is an interdisciplinary one. It must be highlighted that aeronautical field is not tolerable to mistakes. There should not be any mistakes at all. But there is not a certainty in AI either. The most mesmerizing part of reinforced learning is that it makes less mistakes than an actual human pilot. The best demonstration for this claim is the victory of HERON system's AI against a real F-16 pilot in a dogfight. This research proves that aeronautical and artificial intelligence disciplines provides significantly vital outcomes when they collaborate. We are currently working on a research about dogfighting in a 3-dimensional environment, it will be a fundamental research once its done. If we had the change to redo this research, we would not change anything. Because this model proved itself and its success, and it is not changeable when the subject is continuous actions. Aeronautical and artificial intelligence sectors improve day by day, there are many devices that are breakthroughs of technology. So, for the future research subjects, it is possible to see a wider range of possibilities lies within those two disciplines. Currently this research focusing on dogfighting of two aircrafts, paves the way of more advanced UAVs since they are able to perform better in maneuvers, and they are able to learn and perform faster than a human. There are two main advantages for RL, it is inexpensive and easily applicable. Training a pilot takes a long time but also it costs high, with the help of RL, it is possible to integrate this method to many different kinds of aircrafts.

As a result, we first built our model on reaching the goal with a single agent. Then we added it as a barrier to our environment. In Phase 3, we added another agent to our environment and had it do dog-fight. In the last stage, we made a dog-fight with TF-X and F-16, which could be 3 dimension, thanks to behavior cloning.

## A. Appendix

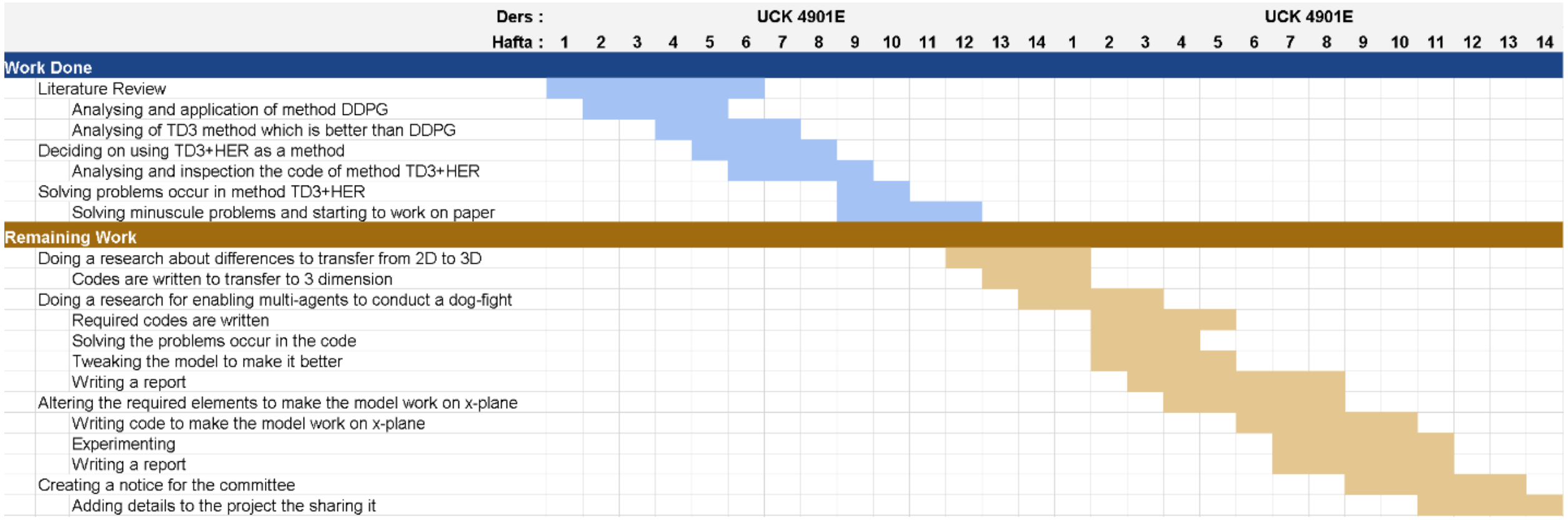

**Figure 18:** **Thesis Schedule.**

## B. References

[1] Kaelbling, Leslie P., Littman, Michael L., Moore, Andrew W., "Reinforcement Learning: A Survey", Journal of Artificial Intelligence Research,Vol.4,1 May, pp. 237-285.
doi: 10.1613/jair.301

[2] Van Otterlo M. and Wiering M., Reinforcement learning and Markov Decision Processes, Springer, Berlin, Heidelberg, 2012,pp. 3-42.
doi.org/10.1007/978-3-642-27645-3_1

[3] S. J. Russell and P. Norvig ,Artificial intelligence: a modern approach, Pearson Education Limited,New-Jersey, 2016,pp.92-115.

[4] R. E. Korf, Artificial intelligence search algorithms,Chapman & Hall/CRC, California,2010,Chap.22.

[5] Kavraki, L. E., Kolountzakis, M. N. and Latombe J., "Analysis of probabilistic roadmaps for path planning," Proceedings of IEEE International Conference on Robotics and Automation, vol.4, 1996, pp. 3020,3025.
doi: 10.1109/ROBOT.1996.509171.

[6] Melchior, N. A. and Simmons, R., "Particle RRT for Path Planning with Uncertainty" ,Proceedings 2007 IEEE International Conference on Robotics and Automation, 2007, pp. 1617,1624.
doi: 10.1109/ROBOT.2007.363555.

[7] Yang, S. X. and Luo, C., "A neural network approach to complete coverage path planning", IEEE Transactions on Systems, Man, and Cybernetics, Part B (Cybernetics), vol. 34, no. 1, 2004, pp. 718,724.
doi: 10.1109/TSMCB.2003.811769.

[8] Malik, M. Z., Eizad , A.,Khan, M. U., Path Planning Algorithms for Mobile Robots, LAP LAMBERT Academic Publishing, Sunnyvale,2014, pp.116.

[9] Alam, M. S., Rafique ,M. U. and Khan ,M. U., "Mobile Robot Path Planning in Static Environments using Particle Swarm Optimization", International Journal of Computer Science and Electronics Engineering (IJCSEE), Vol. 3, No. 3, 2015, pp. 253.257.

[10] M. Khan , "Mobile Robot Navigation Using Reinforcement Learning in Unknown Environments", Balkan Journal of Electrical and Computer Engineering, vol. 7, no. 3, pp. 235-244, Jul. 2019.
doi:10.17694/bajece.532746

[11] Fujimoto, S., Hoof, H. and Meger, D.,"Addressing function approximation error in actor-critic methods", International Conference on Machine Learning , 2018, pp.1587-1596.

[12] Lillicrap, T. P., Hunt, J. J., Pritzel, A., Heess, N., Erez, T., Tassa, Y., Silver, D. and Wierstra, D., "Continuous control with deep reinforcement learning", International Conference on Learning Representations, 2016.

[13] Andrychowicz, M., Filip, W., Alex, R., Jonas, S., Rachel, F., Peter, W., Bob, M., Josh, T., Pieter, A., and Wojciech, Z., "Hindsight experience replay." , Advances in Neural Information Processing Systems, 2017.

[14] Gao, J., Ye, W., Guo, J., Li, Z., "Deep Reinforcement Learning for Indoor Mobile Robot Path Planning", Sensors , Vol. *20*, pp. 5493.
doi.org/10.3390/s20195493